\documentclass[conference]{IEEEtran}
\IEEEoverridecommandlockouts
\usepackage{cite}
\usepackage{amsmath,amssymb,amsfonts}
\usepackage{algorithmic}
\usepackage{graphicx}
\usepackage{textcomp}
\usepackage{xcolor}
\usepackage{hyperref}
\usepackage{multirow}

\def\BibTeX{{\rm B\kern-.05em{\sc i\kern-.025em b}\kern-.08em
    T\kern-.1667em\lower.7ex\hbox{E}\kern-.125emX}}
\begin{document}

\title{Consensus Focus for Object Detection and Minority Classes}

\author{\IEEEauthorblockN{1\textsuperscript{st} Erik Isai Valle Salgado}
\IEEEauthorblockA{\textit{Tsinghua-Berkeley Shenzhen Institute} \\
\textit{Tsinghua University}\\
Guangdong, China \\
gal20@tsinghua.org.cn}
\and
\IEEEauthorblockN{2\textsuperscript{nd} Chen Li}
\IEEEauthorblockA{\textit{Shenzhen International Graduate School} \\
\textit{Tsinghua University}\\
Guangdong, China}
\and
\IEEEauthorblockN{3\textsuperscript{rd} Yaqi Han}
\IEEEauthorblockA{\textit{Shenzhen International Graduate School} \\
\textit{Tsinghua University}\\
Guangdong, China}
\and
\IEEEauthorblockN{4\textsuperscript{th} Linchao Shi}
\IEEEauthorblockA{\textit{Beijing Institute of Technology} \\
\textit{Beijing, China}}
\and
\IEEEauthorblockN{5\textsuperscript{th} Xinghui Li}
\IEEEauthorblockA{\textit{Tsinghua-Berkeley Shenzhen Institute} \\
\textit{Tsinghua University}\\
Guangdong, China}
\and
\and
}

\maketitle

\begin{abstract}
    Ensemble methods exploit the availability of a given number of classifiers or 
    detectors trained in single or multiple source domains and tasks to address machine 
    learning problems such as domain adaptation or multi-source transfer learning.
    Existing research measures the domain distance between the sources and the target dataset, 
    trains multiple networks on the same data with different samples per class, or combines 
    predictions from models trained under varied hyperparameters and settings. Their solutions 
    enhanced the performance on small or tail categories but hurt the rest. To this end, we 
    propose a modified consensus focus for semi-supervised and long-tailed object detection. 
    We introduce a voting system based on source confidence that spots the contribution of 
    each model in a consensus, lets the user choose the relevance of each class in the target 
    label space so that it relaxes minority bounding boxes suppression, and combines multiple 
    models' results without discarding the poisonous networks. Our tests on synthetic driving 
    datasets retrieved higher confidence and more accurate bounding boxes than the NMS, 
    soft-NMS, and WBF. The code used to generate the results is available in our \href{http://github.com/ErikValle/Consensus-focus-for-object-detection}{GitHub repository}.
\end{abstract}


\section{Introduction}
Ensemble techniques make use of a selection of classifiers or detectors 
trained across one or more domains and tasks $(\mathbb{X}^{(i)}, P(\mathbb{X}^{(i)}))$ 
to tackle machine learning challenges. Nowadays, its usage is not limited to 
selecting a classifier but fusing multiple classifiers by training the models 
in local neighborhoods or the whole feature space and combining them to get 
a composite classifier. To this end, we aim to utilize any existing available 
source domains and tasks such that they produce more accurate results for a 
target with some or no labels such that they overcome the following drawbacks:

\begin{enumerate}
    \item \textit{Every source dataset has a different data imbalance rate and may 
    contribute more or less to specific classes}: Hence, collecting inferences from 
    a group of datasets and relaxing the minority bounding boxes filtering out 
    complements the data scarcity in tail categories.
    
    \item \textit{In semi-supervised learning, assuming that the label spaces 
    or domains are identical may imply omitting some relevant entities in the target 
    dataset that could be unknown}. Thus, letting choose the classes in the target 
    label space brings control of what to spot according to the application.
    
    \item \textit{Using discrepant source domains concerning the target distribution 
    or level space may cause negative transfer learning, hurting the overall performance}. 
    Then, a voting system based on source confidence spots their contribution.
\end{enumerate}

\section{Related works}

Given multiple datasets, evaluating their relevance based on how they transfer 
parameters to share knowledge with a target model or among source networks or 
combining their inferences in a voting method for object detection and classification 
is a novel application of ensemble methods. The MJWDEL~\cite{1}. learns transferred 
weights for evaluating the importance of each source set to the target task by 
training sub-models for each source and the target dataset. Then, an attention scheme 
based on the joint Wasserstein distance between the sources and the target domains 
performs the knowledge transference. Finally, the algorithm forms an ensemble model 
by reweighting each sub-model. Since detecting some objects is unsuitable due to their 
size, dataset long-tailedness, or the network training conditions, Chen Li et al.~\cite{2}
suggested training three transfer models with weighted emphasis on minority classes. 
After obtaining the results, a standard threshold and an NMS filter the bounding boxes 
from the mentioned networks. This solution enhances tail category performance but hurts 
metrics for the rest. Focusing on optimized training for specific classes, Enrique 
Daherne et al.~\cite{3} employed Weighted Box Fusion to combine predictions from models 
trained under varied hyperparameters and settings. Their findings only focused on 
hyperparameter tuning and performance assessment, which does not necessarily fit other 
applications. Roman Solovyev et al.~\cite{4} presented an ensemble method that combines 
predictions from multiple detectors, comparing their proposed Weighted Box Fusion against 
varied NMS methods. Assigning weights to define which model benefits the ensemble could 
be intricate if the target dataset distribution or task is unknown. Thus, we propose a 
modified consensus focus as a dynamic weighting strategy for object detection to compute 
each model's contribution.

\section{Approach}
First, we obtain the inferences $f^{(i)}(x_j^{(T)})=B_j^{(i)}$ of the source domain 
models on each unlabeled target domain data $x_j^{(T)}\in X^{(T)}$. Next, we utilize 
the WBF to get the bounding-box consensus knowledge $B_j$ of the source networks. 
Finally, we also set an extended source dataset $D^{(I+1)}$ with the bounding box
consensus knowledge for each target domain image $x_j^{(T)}$ that works for model 
ensembling and domain adaptation purposes after the weighted model aggregation.

\begin{align}
    D^{(I+1)}=\left\{\left(x_j^{\left(T\right)},B_j\ \right)\right\}_{j=1}^{M^{(T)}} \label{eq1}
\end{align}
    
where $B_j={b_1,..,b_j}$ are the predicted bounding boxes for $x_j^{\left(T\right)}$ 
after applying the WBF (each includes the category, bounding box coordinates, 
confidence, and the number of domains $n_{b_j}$ that supports the prediction), and $M^{(T)}$ 
is the number of images in the target domain. In the end, the contribution $\alpha_i^{CF}$ 
that every source model has over the target domain derived from the quality of consensus 
lets the user obtain source-weighted inferences and sharpen bounding boxes with trustable 
confidence, even feeding the group of datasets $D$ with an unrelated source. The next 
subsections will describe the steps in detail.

\subsection{Knowledge Vote for Object Detection and Minority Classes}
The Knowledge Vote for Object Detection determines the most likely true bounding boxes 
supported by more source domains with high confidence through the certain consensus 
knowledge defined as follows:
    
\begin{enumerate}
    \item Delimiting the target label space. The users can select the categories to 
    filter based on their experience, the relevance a set of defects has, or the data 
    imbalance trend the source datasets have. To this end, there are two alternatives: 
    removing the chosen labeled bounding boxes from every source dataset or assigning 
    a high confidence gate $c_g=\left\{c_{g_1},\ldots,c_{g_\kappa}\right\}$ 
    to each category $\kappa$ accordingly. If the target label space is undefined, 
    the method will consider the same value for every class in the global label space. 
    \item After filtering out the non-relevant labels, the WBF merges those bounding 
    boxes by following the steps described in~\cite{4} but counting the number of models 
    $n_{b_j}$ that formed the final box coordinates and its confidence score $p_{b_j}$. 
    The last parameter let us measure the contribution of each network per instance 
    $x_j^{(T)}$ during consensus-focus computing.
\end{enumerate}   

\begin{figure}[htbp]
    \centerline{\includegraphics[width=\linewidth]{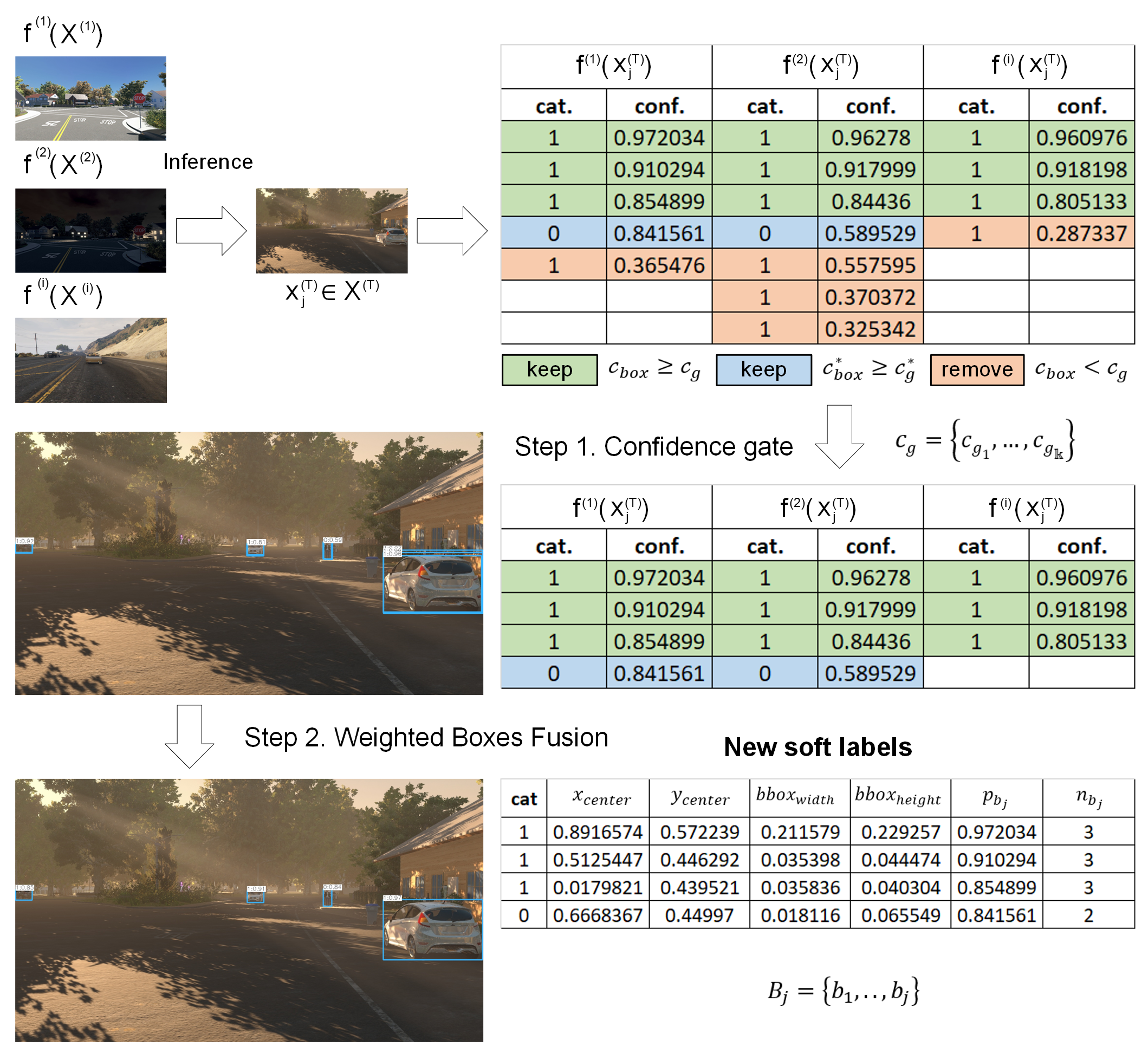}}
    \caption{Knowledge vote ensemble for object detection.}
    \label{fig1}
\end{figure}

Fig.~\ref{fig1} shows an example of all the bounding boxes inferred by three networks 
trained over different source datasets. In that case, the tailed classes (category 1 
corresponds to pedestrians) got a tolerant confidence gate ($c_g^\ast\geq0.5$), while 
the rest was ($c_g\geq0.8$). Next, the WBF algorithm sorted and merged them accordingly, 
where the first box achieved the highest confidence and more models supported it.

\subsection{Consensus Focus for Object Detection}
The purpose of Consensus Focus is to spot the divergent domains that lead to negative 
transfer or undesired inferences and re-weight each source domain to maximize its 
contribution to the target domain. Other authors~\cite{5, 6} proposed methods for measuring 
the domain discrepancy on the input space. However, they rely on the amount of data every 
distribution has or a probability discrepancy measure, analyzing only the distribution 
rather than the label information provided. We can confirm that only measuring the similarity 
among domains will likely fail to identify poisonous datasets.

Under the premise of preserving data privacy, we can adjust the definition of consensus quality 
for object detection as the sum of the product between the number of models supporting a fused 
bounding box in a source dataset subset $S^\prime\subseteq S$. Here, we denote $S=\left\{D^{\left(i\right)}\right\}_{i=1}^I$ 
as a set of source datasets and $Q\left(S^\prime\right)$ is the consensus quality of a subset of $S$:

\begin{align}
Q\left(S^\prime\right)=\sum_{x_j^{\left(T\right)}\in X^{\left(T\right)}}\sum_{b_j\in B_j}{n_{b_j}\left(S^\prime\right)}p_{b_j}(S^\prime) \label{eq2}
\end{align}

The consensus focus $CF\left(D^{\left(i\right)}\right)$ quantifies the contribution of each source 
domain through the surveys based on measuring the total consensus quality of all combinations 
between the elements of $S$. In other words, the $CF$ designates the marginal contribution of the 
single source domain $D^{\left(i\right)}$ to the consensus quality of all source domains $S$.

\begin{align}
CF\left(D^{\left(i\right)}\right)=Q\left(S\right)-Q\left(S\setminus\left\{D^{\left(i\right)}\right\}\right) \label{eq3}
\end{align}

To compute the dataset weights given the new source domain $D^{\left(I+1\right)}$, we first 
calculate the weight $\alpha_{I+1}^{CF}$.

\begin{align}
    \alpha_{I+1}=\frac{M^{\left(T\right)}}{M^{\left(T\right)}+\sum_{i=1}^{I}M^{(i)}} \label{eq4}
\end{align}

Lastly, $\alpha_i$ is the re-weighting term normalized for each source domain noted as
\begin{align}
    \alpha_i^{CF}=\left(1-a_{I+1}\right)\frac{M^{\left(i\right)}CF\left(D^{\left(i\right)}\right)}{\sum_{i=1}^{I}{M^{\left(i\right)}CF\left(D^{\left(i\right)}\right)}} \label{eq5}
\end{align}

In~\cite{7}, the publication utilized the previous expression to adjust the training parameters 
of the target model by adding the $I+1$ source networks through a Consensus Focus for object 
classification. In our case, the calculated weights amend the initial parameters used in 
the WBF to bring the highest confidence based on the contribution of each source dataset. 
Indeed, our approach also can evaluate the source domains in a semi-supervised setting via 
knowledge distillation or even in federated learning applications.

\section{Results and Discussion}
We considered three synthetic datasets for autonomous driving: Apollo Synthetic~\cite{8}, FCAV~\cite{9}, 
and Virtual KITTY 2~\cite{10}. Since the first has more than 273k distinct images, our tests take 
two out of seven parts (13-00 and 18-00) with clear and heavy rain, all degradations, pedestrians, 
traffic barriers, and all scenes. The target dataset belongs to an Apollo subset created by 
simulating the daytime at 5 pm with a clear sky and including the previous environmental variations. 
Each target class has 1333, 4556, and 234 samples, accordingly. For this experiment, we group all 
the categories into three classes to homogenize the label space: pedestrian, motorized vehicle 
(e.g., car, pickup, truck, etc.), and non-motorized-vehicle (cyclist, motorcyclist, unicyclist, etc.). 
YOLOv8x~\cite{11} is the benchmark model to compare the performance of our approach with the NMS, 
soft-NMS, and WBF.

\begin{figure}[htbp]
    \centerline{\includegraphics[width=\linewidth]{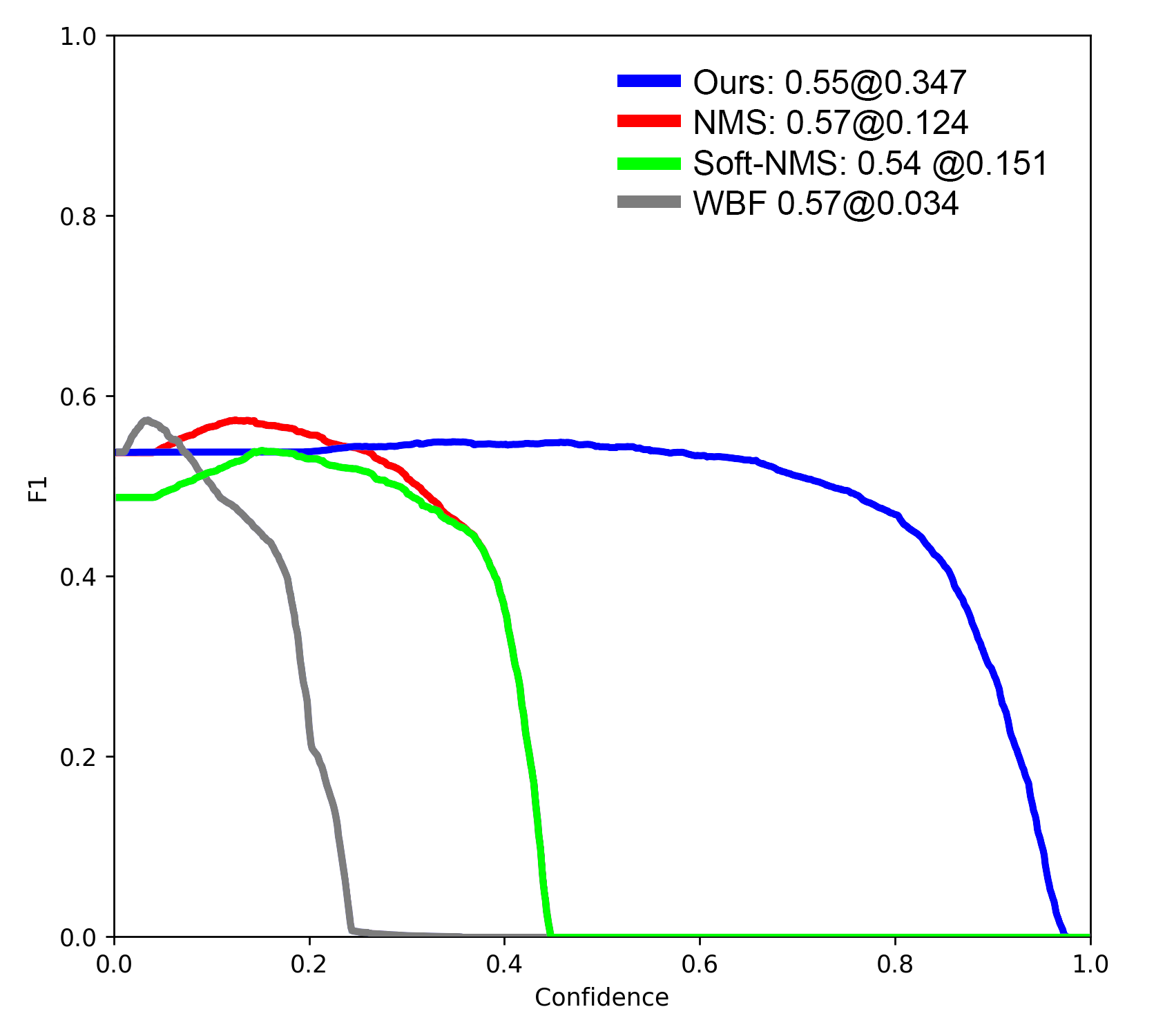}}
    \caption{F1 curves comparing the performance of our method versus NMS, 
    Soft-NMS, and WBF (all weights equal to one) after skipping boxes with 
    confidence lower than 0.0001.}
    \label{fig2}
\end{figure}

Although the NMS metrics in Table~\ref{tab1} for boxes with a confidence threshold above 0.0001 
are slightly higher than our approach, the F1 curve in Fig.~\ref{fig2} shows how other 
methods reach their maximum F1 score before a confidence value of 0.5. In contrast, 
if we raise such a threshold to 0.3, the precision will increase as the other's mAPs 
drop drastically, suggesting that the robustness of our solution overcomes the rest 
when the confidence is crucial. Indeed, its usage is suitable for appliances with 
limited labeled target data (e.g., semi-supervised learning).

%tab1
\begin{table*}[h!t]
    \center
    \caption{Comparison of our method vs others}
    \begin{tabular}{llllll}
    \hline
    Confidence threshold & Ensemble method & P & R & mAP@0.5 & mAP@.5:.95 \\
    \hline
    \multirow{4}{4em}{0.0001} & Ours & 0.775 & 0.432 & 0.43 & 0.313\\
    & NMS & 0.856 & 0.435 & 0.434 & 0.322\\
    & Soft-NMS & 0.748 & 0.426 & 0.421 & 0.316\\
    & WBF & 0.868 & 0.432 & 0.435 & 0.316\\
    \hline
    \multirow{4}{4em}{0.0003} & Ours & 0.779 & 0.431 & 0.425 & 0.311\\
    & NMS & - & - & - & - \\
    & Soft-NMS & 0.806 & 0.409 & 0.407 & 0.303\\
    & WBF & 0.726 & 0.439 & 0.421 & 0.301\\
    \hline
    \end{tabular}
    \label{tab1}
\end{table*}

\section{Conclusions}
In this work, we introduced a two-step object-detection consensus focus. First, it 
removes bounding boxes failing the class-oriented confidence gates to ensure the prediction 
quality, and then the WBF merges the remaining detections. Next, the modified consensus 
focus measures the consensus quality of each combination of source model predictions to 
estimate their contribution, assigning a weight to them so that the final inference relaxes 
the relevance of the poisonous domains. The results suggest that our method retrieves higher 
confidence and more accurate bounding boxes than the NMS, soft-NMS, and WBF. Nevertheless, 
it comes at a cost of increased processing time—approximately three times that of standard 
NMS—due to the heuristic process involved in the combinatorial analysis for consensus focus. 
In the future, we plan to implement this technique for federated learning.

\section{Code, Data, and Materials Availability}
All data in support of the findings of this paper are available within the article or as supplementary material.
Thus, the code used to generate the results and figures is available in a Github repository at \url{http://github.com/ErikValle/Consensus-focus-for-object-detection}.

\end{document}